\documentclass{article}

\PassOptionsToPackage{numbers, compress}{natbib}

\usepackage[final]{neurips_2021}

\usepackage[utf8]{inputenc} 
\usepackage[T1]{fontenc}    
\usepackage{hyperref}       
\usepackage{url}            

\usepackage{booktabs}       
\usepackage{amsfonts}       
\usepackage{nicefrac}       
\usepackage{microtype}      
\usepackage{xcolor}         

\usepackage{amsmath}
\usepackage{amssymb}

\title{Medical Codes Prediction from Clinical Notes: \\ From Human Coders to Machines}

\author{%
  Byung-Hak Kim\\
  AKASA, Inc.\\
  South San Francisco, CA\\
  \texttt{hak.kim@AKASA.com} \\
}

\begin{document}
\maketitle
\section{Introduction}
Prediction of medical codes from clinical notes is a practical and essential need for every healthcare delivery organization within current medical systems\footnote{A human coder or health care provider scans the medical documentation in electronic health records, identifying essential information and annotating codes for that particular treatment or service. With a wide range of medical services and providers (primary care clinics, specialty clinics, emergency departments, mother-baby units, outpatient and inpatient units, etc.), the complexity of human coders' tasks increases as the medical industry advances, while productivity standards decrease as charts take more time to review. }. Automating annotation will save significant time and excessive effort that human coders spend today. However, the biggest challenge is directly identifying appropriate medical codes from several thousands of high-dimensional codes from unstructured free-text clinical notes. This complex medical codes prediction problem from clinical notes has received substantial interest in the NLP community, and several recent studies~\citep{Kim21,Sun21,Liu21,Yuan22} have shown the state-of-the-art (SOTA) code prediction results of full-fledged deep learning-based methods. This progress raises the fundamental question of how far automated machine learning (ML) systems are from human coders' working performance, as well as the important question of how well current explainability methods apply to advanced neural network models such as transformers. This is to predict correct codes and present references in clinical notes that support code prediction, as this level of explainability and accuracy of the prediction outcomes is critical to gaining trust from professional medical coders. 

\section{Explainable and Accurate Medical Codes Prediction}
\textbf{RAC Model and Human-Level Coding Baselines:} We first present our \textbf{R}ead, \textbf{A}ttend, and \textbf{C}ode (RAC) model to learn the assignment mappings of medical codes that can process unstructured clinical notes and attend to text areas annotating medical codes. The main building blocks of RAC are built upon by connecting convolved embeddings with Transformer-encoder based self-attention and code-title guided attention modules. The architecture details can be found in Section 4 of ~\citep{Kim21}. In principle, we look at the medical codes prediction problem as a set-to-set assignment learning problem from the set of input sentence vectors to the set of code labels and employ the problem's unique permutation equivariant property\footnote{Consider permuting the notes' sentence vectors; medical code outputs will be the same but permuted.} in the design. 

Second, the human coding baseline is estimated via a primitive web interface built using the Label Studio~\citep{LS20}. Given the reference diagnosis and procedure codes that Beth Israel Deaconess Medical Center (BIDMC) assigned, we have two professional CPC certified coders independently code the different set of notes and evaluate where their total annotations have differed from the references. The intention is to provide the same possible conditions as for the RAC system. We found that estimated inter-coder agreement rates (i.e., human coding baseline) were not as high as we initially thought and are far exceeded by the RAC system's 3.9 times higher rate in Micro-Jaccard similarity. Section 3 of ~\citep{Kim21} conatins the evaluation design details and baseline results.

Third, based on the MIMIC-III dataset\footnote{The MIMIC-III Dataset (MIMIC v1.4) is a freely accessible medical database containing de-identified medical data of over 40,000 patients staying in the BIDMC between 2001 and 2012.}, we demonstrate the RAC model's effectiveness in the most challenging full codes prediction testing set from \emph{inpatient} clinical notes. The RAC model wins over all the previously reported SOTA results considerably. 
RAC establishes a new SOTA, considerably outperforming the current best Macro-F1 by 18.7\%, and reaches past the human-level coding baseline. For further information on prediction results, see Table 1 and 2 of ~\citep{Kim21}, as well as the Papers with Code's leaderboard in the medical code prediction~\citep{PWC22}. 

\textbf{xRAC Framework and Human-Grounded Explainability Evaluation:} Next, we present a general e\textbf{x}plainable \textbf{R}ead, \textbf{A}ttend, and \textbf{C}ode (xRAC) framework that generates evidentiary text snippets for a predicted code, which are oriented towards the needs of a deployment scenario. The first attention score-based approach, xRAC-ATTN, utilizes the label-wise attention mechanism first introduced in~\citep{Mullenbach18} to select key sentences for prediction decisions.
In the second model-agnostic knowledge-distillation-based method, xRAC-KD, a large "teacher" RAC model is distilled into a collection of "student" linear models in post-hoc manner without sacrificing much accuracy of the teacher model while retaining many advantages of linear models including explainability and smaller model size which is beneficial for deployment. More methodological details are available in Section 2 of~\citep{Kim22}. 

Then, a simplified but thorough human-grounded evaluation with two groups of annotators, one group (Group A) with and one group without (Group B) medical coding expertise\footnote{Group A had two annotators without medical coding experience and Group B had six certified professional coders.}, is conducted to evaluate the explainability of our xRAC framework, xRAC-ATTN and xRAC-KD. Both groups are given the same question sheet, which has explanation text snippets extracted to support the appearance of the predicted codes. Each annotator must select one of the three choices that are highly informative, informative, and irrelevant. More information on the evaluation task design and execution can be found in Section 3.2 of~\citep{Kim22}. 

We find that the proposed xRAC framework can be easily explainable and the supporting evidence text highlighted by xRAC-ATTN is of higher quality than xRAC-KD even though xRAC-KD has potential advantages in production deployment scenarios. One can see from Table 2 in~\citep{Kim22} that there is much larger gap in xRAC-KD between Group A and Group B than between xRAC-ATTN. This implies that xRAC-ATTN is a more viable choice than xRAC-KD to extract a text snippet from clinical notes to support code prediction. However, Table 3 in~\citep{Kim22} shows that the consistency score measured by Jaccard Similarity between two groups is lower than 40\% even with xRAC-ATTN. 

This suggests that the automated extraction system must still rely on professional coders' feedback, and there is room remaining to improve for a lay person without expertise to correctly code. We believe that this is a very meaningful step toward the goal of developing accurate, explainable and automated ML systems for medical code prediction from clinical notes. To our knowledge, most previous works are still in early stages in terms of providing textual references and explanations of the predicted codes and no studies to date have thoroughly analyzed in this regard, especially for complex transformer-based models such as the RAC model.

\section{Conclusion}
In this study, we present for the first time a human-coding baseline for medical code prediction on the subsampled MIMIC-III-full-label inpatient clinical notes testing set task. We have developed an attention-based RAC model that sets the new SOTA records, and the resulting RAC model outperforms the human-coding baseline to a great extent on the same task. The performance improvements can be attributed to effectively learning the common embedding space between the clinical note and medical codes by utilizing attention mechanisms that efficiently address the severe long-tail sparsity issues. Additionally, we present a xRAC framework to obtain supporting evidence text from clinical notes that justify predicted medical codes from medical code prediction systems. We have demonstrated that the proposed xRAC framework may help even complex transformer-based models to attain high accuracy with a decent level of explainability (which is of high value for deployment scenarios) through qualitative human-grounded evaluations. We also show for the first time that, given the current state of explainability methodologies, using the proposed explainable yet accurate medical codes prediction system still requires professional coders' expertise and competencies.

\bibliography{neurips_2021}

\begin{thebibliography}{8}
\providecommand{\natexlab}[1]{#1}
\providecommand{\url}[1]{\texttt{#1}}
\expandafter\ifx\csname urlstyle\endcsname\relax
  \providecommand{\doi}[1]{doi: #1}\else
  \providecommand{\doi}{doi: \begingroup \urlstyle{rm}\Url}\fi

\bibitem[Kim and Ganapathi(2021)]{Kim21}
B.-H. Kim and V.~Ganapathi.
\newblock Read, {A}ttend, and {C}ode: Pushing the limits of medical codes
  prediction from clinical notes by machines.
\newblock In \emph{Proceedings of the 6th Machine Learning for Healthcare
  Conference (MLHC 2021)}, volume 149 of \emph{Proceedings of Machine Learning
  Research}, pages 196--208. PMLR, 2021.

\bibitem[Kim et~al.(2022)Kim, Deng, Yu, and Ganapathi]{Kim22}
B.-H. Kim, Z.~Deng, P.~Yu, and V.~Ganapathi.
\newblock Can current explainability help provide references in clinical notes
  to support humans annotate medical codes?
\newblock In \emph{Proceedings of the 13th International Workshop on Health
  Text Mining and Information Analysis (Louhi 2022)}. Association for
  Computational Linguistics, 2022.

\bibitem[Liu et~al.(2021)Liu, Cheng, Klopfer, Gormley, and Schaaf]{Liu21}
Y.~Liu, H.~Cheng, R.~Klopfer, M.~R. Gormley, and T.~Schaaf.
\newblock Effective convolutional attention network for multi-label clinical
  document classification.
\newblock In \emph{Proceedings of the 2021 Conference on Empirical Methods in
  Natural Language Processing (EMNLP 2021)}, pages 5941--5953, Online and Punta
  Cana, Dominican Republic, 2021. Association for Computational Linguistics.

\bibitem[Mullenbach et~al.(2018)Mullenbach, Wiegreffe, Duke, Sun, and
  Eisenstein]{Mullenbach18}
J.~Mullenbach, S.~Wiegreffe, J.~Duke, J.~Sun, and J.~Eisenstein.
\newblock Explainable prediction of medical codes from clinical text.
\newblock In \emph{Proceedings of the 2018 Conference of the North American
  Chapter of the Association for Computational Linguistics: Human Language
  Technologies, Volume 1 (Long Papers)}, page 1101–1111, New Orleans,
  Louisiana, 2018. Association for Computational Linguistics.

\bibitem[{Papers With Code: The latest in Machine Learning}(2022)]{PWC22}
{Papers With Code: The latest in Machine Learning}.
\newblock {Medical Code Prediction on MIMIC-III}, 2022.
\newblock URL
  \url{https://paperswithcode.com/sota/medical-code-prediction-on-mimic-iii?metric=Macro-F1}.

\bibitem[Sun et~al.(2021)Sun, Ji, Cambria, and Marttinen]{Sun21}
W.~Sun, S.~Ji, E.~Cambria, and P.~Marttinen.
\newblock Multitask balanced and recalibrated network for medical code
  prediction.
\newblock In \emph{European Conference on Machine Learning and Principles and
  Practice of Knowledge Discovery in Databases (ECML PKDD 2021)}, 2021.

\bibitem[Tkachenko et~al.(2020)Tkachenko, Malyuk, Shevchenko, Holmanyuk, and
  Liubimov]{LS20}
M.~Tkachenko, M.~Malyuk, N.~Shevchenko, A.~Holmanyuk, and N.~Liubimov.
\newblock {Label Studio}: Data labeling software, 2020.
\newblock URL \url{https://github.com/heartexlabs/label-studio}.
\newblock Open source software available from
  https://github.com/heartexlabs/label-studio.

\bibitem[Yuan et~al.(2022)Yuan, Tan, and Huang]{Yuan22}
Z.~Yuan, C.~Tan, and S.~Huang.
\newblock Code {S}ynonyms {D}o {M}atter: Multiple synonyms matching network for
  automatic icd coding.
\newblock In \emph{Proceedings of the 60th Annual Meeting of the Association
  for Computational Linguistics (ACL 2022)}, 2022.

\end{thebibliography}

\end{document}